\documentclass[conference]{IEEEtran}
\usepackage{times}

\usepackage{algorithm}
\usepackage[noend]{algpseudocode}
\usepackage{amsmath}
\usepackage{amssymb}
\usepackage{amsthm}
\usepackage{booktabs}
\usepackage{caption}
\usepackage{color}
\usepackage[bookmarks=true,colorlinks=true,citecolor=blue,linkcolor=blue]{hyperref}
\let\labelindent\relax 
\usepackage{enumitem}
\usepackage{graphicx}
\usepackage{makecell}
\usepackage{mathtools}
\usepackage{microtype}
\usepackage{multirow}
\usepackage{multicol}
\usepackage{subcaption}
\usepackage{tabularx}
\usepackage{tikz}
\usetikzlibrary{positioning,arrows.meta,fit,calc,backgrounds,decorations.pathmorphing}
\usepackage{tikzscale}
\usepackage{url}
\usepackage{wrapfig}
\usepackage{xcolor}
\usepackage{xparse}
\usepackage{xspace}
\usepackage{xpatch}

\usepackage{subfiles}
\usepackage[textsize=scriptsize]{todonotes} 
\IEEEoverridecommandlockouts

\makeatletter
\let\NAT@parse\undefined
\makeatother

\makeatletter
\def\mathcolor#1#{\@mathcolor{#1}}
\def\@mathcolor#1#2#3{%
  \protect\leavevmode
  \begingroup
    \color#1{#2}#3%
  \endgroup
}
\makeatother

\newcommand{\srceq}[1]{\mathcolor{purple}{#1}}
\newcommand{\tgteq}[1]{\mathcolor{blue!60}{#1}}

\usepackage[numbers,sort&compress]{natbib}
\bibpunct{[}{]}{,}{n}{}{;}

\theoremstyle{plain}

\theoremstyle{definition}

\theoremstyle{remark}

\usepackage{amsmath,amsfonts,bm}

\def\eqref#1{equation~\ref{#1}}

\def\1{\bm{1}}

\DeclareMathAlphabet{\mathsfit}{\encodingdefault}{\sfdefault}{m}{sl}
\SetMathAlphabet{\mathsfit}{bold}{\encodingdefault}{\sfdefault}{bx}{n}

\DeclareMathOperator*{\argmin}{arg\,min}

\usepackage{acronym}
\acrodef{ILA}{invariance through latent alignment}

\newcommand{\ila}{Invariance through Latent Alignment\xspace}
\newcommand{\lcila}{invariance through latent alignment}
\newcommand{\src}{\textrm{src}}
\newcommand{\csrc}{\text{\color{purple}src}}
\newcommand{\tgt}{\textrm{tgt}}
\newcommand{\ctgt}{\text{\color{blue!60}tgt}}

\makeatletter
\newcommand{\printfnsymbol}[1]{%
  \textsuperscript{\@fnsymbol{#1}}%
}
\makeatother

\definecolor{mydarkblue}{rgb}{0,0.08,0.45}
\definecolor{es-blue}{rgb}{0,0.4,0.8}
\definecolor{darkgreen}{rgb}{0.0, 0.5, 0.0}
\definecolor{arsenic}{rgb}{0.23, 0.27, 0.29}

\begin{document}

\title{Invariance Through Latent Alignment}

\author{Takuma Yoneda\textsuperscript{*\printfnsymbol{4}}, 
Ge Yang\textsuperscript{*\printfnsymbol{2}\printfnsymbol{3}},  Matthew R.\ Walter\textsuperscript{\printfnsymbol{4}}, Bradly C.\ Stadie\textsuperscript{\printfnsymbol{4}}
\vspace{2mm}\\

\printfnsymbol{4}Toyota Technological Institute at Chicago (TTIC) \\
\printfnsymbol{2}Institute of Artificial Intelligence and Fundamental Interactions (IAIFI) \\
\printfnsymbol{3}Computer Science and Artificial Intelligence Laboratory (CSAIL), MIT \\

%
%
\texttt{%
\href{mailto:takuma@ttic.edu}{\color{black}takuma@ttic.edu},%
\href{mailto:geyang@csail.mit.edu}{\color{black}geyang@csail.mit.edu}
}
}






\maketitle
\begin{abstract}
A robot's deployment environment often involves perceptual changes that differ from what it has experienced during training. 
Standard practices such as data augmentation attempt to bridge this gap by augmenting source images in an effort to extend the support of the training distribution to better cover what the agent might experience at test time. In many cases, however, it is impossible to know test-time distribution-shift \textit{a priori}, making these schemes infeasible. In this paper, we introduce a general approach, called \textbf{\underline I}nvariance through \textbf{\underline L}atent \textbf{\underline A}lignment (\textbf{ILA}), that improves the test-time performance of a visuomotor control policy in deployment environments with unknown perceptual variations. ILA performs unsupervised adaptation at deployment-time by matching the distribution of latent features on the target domain to the agent's prior experience, without relying on paired data. Although simple, we show that this idea leads to surprising improvements on a variety of challenging adaptation scenarios, including changes in lighting conditions, the content in the scene, and camera poses. We present results on calibrated control benchmarks in simulation---the distractor control suite---and a physical robot under a sim-to-real setup. Video and code available at:~\url{https://invariance-through-latent-alignment.github.io}
\end{abstract}
\def\thefootnote{*}\footnotetext{Equal contribution.}\def\thefootnote{\arabic{footnote}}

\section{Introduction}

Reinforcement learning for control has achieved great success in a wide variety of challenging sensory-motor control tasks, including agile drone flight~\cite{Kaufmann2018drone,Kaufmann2020drone,Loquercio2021drone}, deformable object manipulation~\cite{Wu2019deformable}, and quadruped locomotion~\cite{Hwangbo2019agile,Miki2022wild,Lee2020terrain,margonlis2022rapid}. In comparison to their classical model-predictive control counterparts, reinforcement learning-based approaches enables the use of more realistic forward dynamics model in the form of a physics simulator. Improvements in rigid-body simulator technologies~\cite{Todorov2012mujoco, Makoviychuk2021Isaac} allows reinforcement learning algorithms to overcome their prohibitively-high sample complexity by first training in simulation and then deploying directly on the physical robot. 
Differences, however, still exist between what the controller experiences in the simulator and in the physical environment in the form of a \textit{sim-to-real gap}. In particular, the ability to produce visuomotor control policies that remain robust when perceptual conditions change during deployment, remains an open problem.

\begin{figure}[t!]
    \centering
    \includegraphics[width=\linewidth]{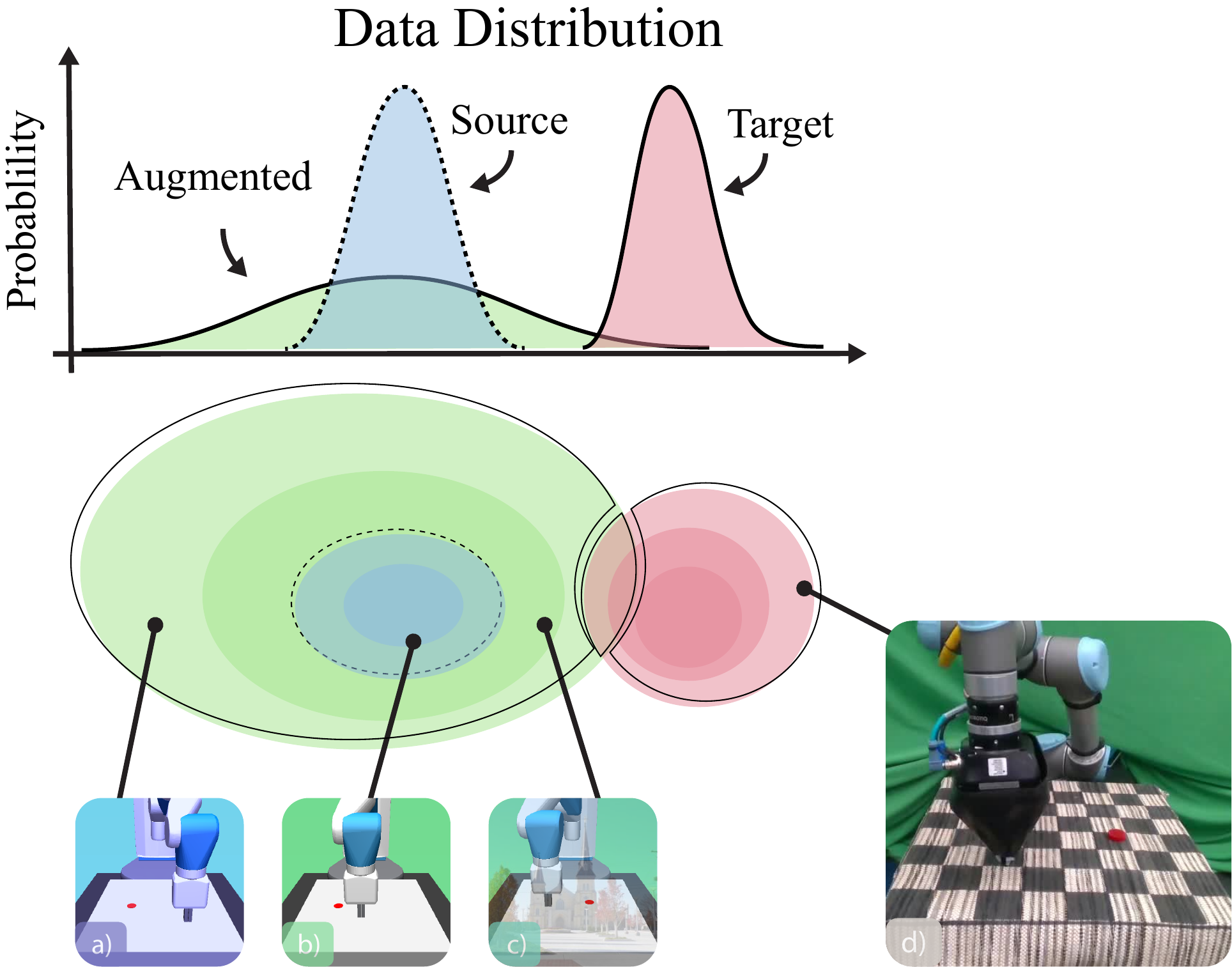}%
    \caption{Data augmentation improves the coverage of the training distribution at the expense of learning complexity and performance. 
    Insets: (a, c) augmented training data; (b) the original input image from the source domain; 
    and (d) the input image in the target domain. Data augmentation fails to provide sufficient coverage of the unknown deployment condition, so the learned controller fails to accomplish the task.}
    \label{fig:out-of-distribution}
\vspace{-1em}
\end{figure}
Consider the illustration in Figure~\ref{fig:out-of-distribution}. A common approach to battle domain shift is to expose the agent to a large variety of data during training with the hope that the training distribution provides adequate coverage over what the agent will experience in the wild. When the simulator is extensible, one can use domain randomization to generate more diverse training data~\cite{tobin2017domain,active-domain-randomization}. Alternatively, one can use data augmentation mechanisms to decorate existing data~\cite{yarats2021drqv2,hansen2020deployment}. Both approaches aim to produce visual features that are invariant to perceptual changes orthogonal to the task. Such invariance does not come for free, however. Additional training slows down the wall-clock speed of the training process, while domain randomization requires manual tuning~\cite{Andrychowicz2020dexterous} and relies on the assumption that the policy network has sufficient capacity to handle the increased support of the input distribution. The added complexity can negatively affect model and policy performance~\cite{laskin2020reinforcement,hansen2020deployment}. Further more, hidden beneath is the assumption that one needs to know roughly what types of perceptual shift would occur during deployment. Failure can happen when the target domain is not known \textit{a priori} and falls \textit{out-of-distribution}, resulting in a fumbling robot that is unable to self-correct (see Figure~\ref{fig:out-of-distribution}). 

Different from these prior approaches that produce invariance by memorizing what is irrelevant to the task during training, we consider a more challenging, but also more realistic scenario in which the specifics of the deployment is not known in advance. This requires the agent to truly generalize \textit{out-of-distribution}, without prior knowledge of the target environment. We further assume that reward supervision is unavailable, so fine-tuning via reinforcement learning is out of the question. This might seem to be an impossible task, but it does implicitly make the assumption that the \textit{same task} that the agent was optimized for during training remains well-defined in the target domain. This means that the agent has \textit{some} notion of \textit{what it knows} despite of the sudden appearance of many unknowns that are not required for the task of interest. Without further assumptions or loss of generality, this \textit{out-of-distribution} generalization problem can be formulated as \textit{unsupervised policy adaptation} between two MDPs that share the same latent dynamics and reward structure, but with distinct pixel observations (See Figure~\ref{fig:graphicalmodels}). 


\begin{figure}[t!]
    \centering
    \includegraphics[width=\linewidth]{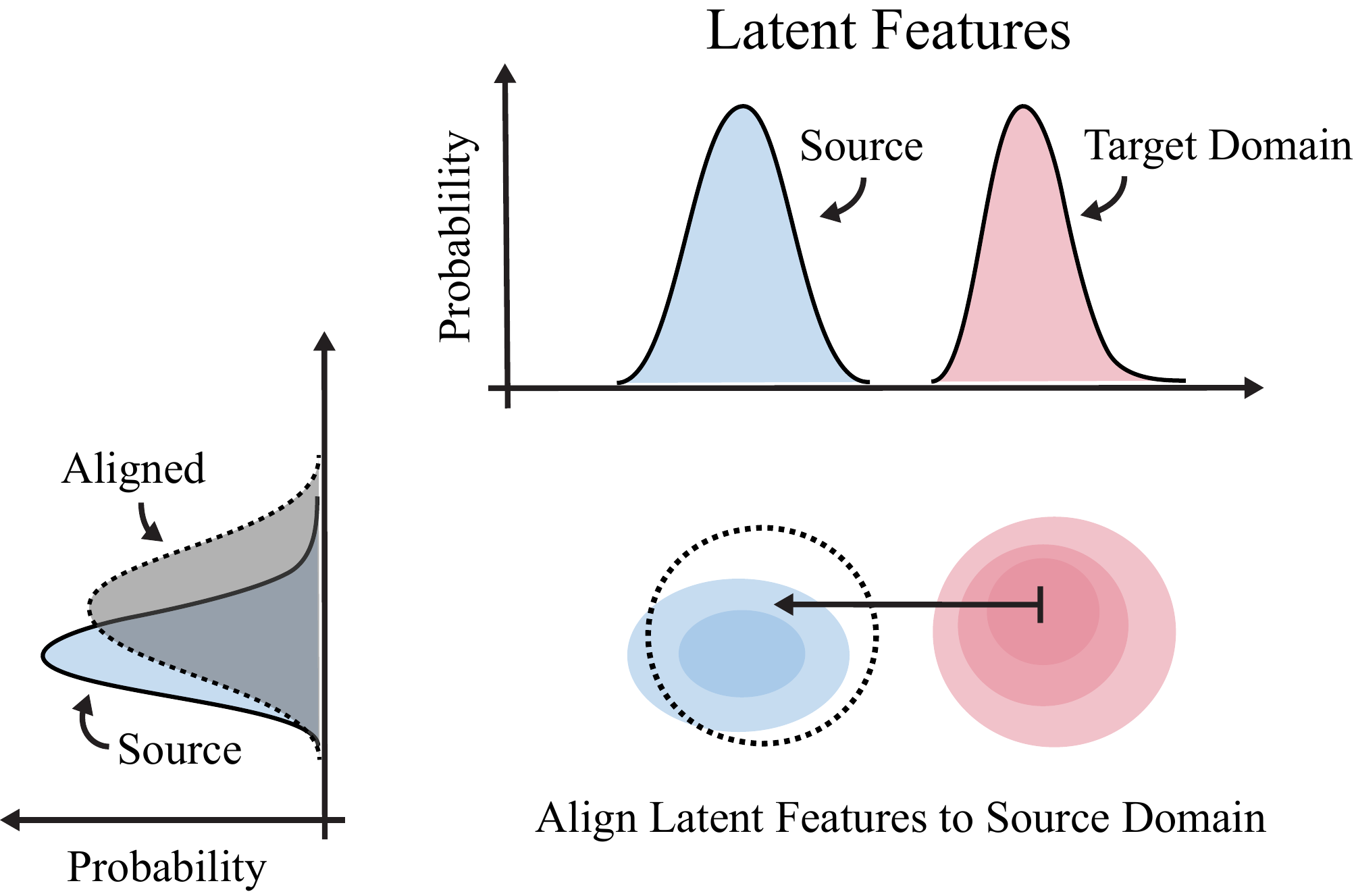}
    \caption{
    Latent features of the network experience a covariant shift in the target domain. Our proposed method, \textbf{\underline I}nvariance through \textbf{\underline L}atent \textbf{\underline A}lignment (\textbf{ILA}), counteracts this domain shift by projecting the features back to the known distribution of latent features on the source domain. We formulate this as a distribution-matching objective in Section~\ref{sec:ila}.
    }\label{fig:latent-alignment}
\vspace{-1em}
\end{figure}
In this paper, we investigate ways to improve generalization under this challenging scenario. Rather than battling domain shift by baking perceptual invariance explicitly into the network during training, we demonstrate a way to produce feature invariance at the time of deployment by taking advantage of the fact that the task of interest remains the same, therefore what the agent experiences internally should remain the same as well. We collect latent features collected during training as examples of what the agent \textit{knows} about the task. On the target domain (see Figure~\ref{fig:latent-alignment}), the new latents are shifted from these prior distributions. Our goal in unsupervised policy adaptation is to match the distribution of these latent features on the target domain with those that appeared during training. This unsupervised learning objective, which we refer to as \textit{latent alignment}, does not require paired image data between the source and the target, and can be applied to any agent without imposing specific requirements of how it is trained. To distinguish our approach from prior works that attempt to produce generalization by baking invariances into the policy at training time, we refer to our method as \textbf{\underline I}nvariance through \textbf{\underline L}atent \textbf{\underline A}lignment (\textbf{ILA}).

\section{Related Work}

A large body of work is dedicated to improving the ability for neural networks to generalize.  These works can largely be placed under two categories---the first category, including domain randomization~\cite{tobin2017domain}, data augmentation~\cite{hansen2021soda}, invariant risk minimization~\cite{Arjovsky2019invariant,Zhang2020invariant}, and meta-learning~\cite{Finn2017maml}, all make the assumption that one has a rough idea of what type of perceptual change is going to occur during deployment. For example, to get the best result with data augmentation, one needs to fine-tune the weights between different augmentation mechanisms because each produces a different type of invariance~\cite{rad2020}. Variations in the camera pose, for instance, is a common problem in robotics. Yet it can not be fixed by augmenting images alone~\cite{stone2021distracting}. Invariance to the projective geometry requires randomizing camera extrinsics during rendering~\cite{hansen2021soda}.

Similarly, meta-learning makes the assumption that one has access to a meta-distribution of task-environment pairs. This is an even stronger assumption than those typically made by supervised learning, which merely requires that training data is sampled from an \textit{i.i.d.}\ that covers the test distribution. In many cases and especially in reinforcement learning, generalization comes from exposure to a large amount of diverse data~\cite{Cobbe2019procgen}. Meta-learning offers little gain procedural-wise, because a meta-learning reinforcement learning algorithm is identical to multi-task training from a task distribution plus fine-tuning that relies on rewards being available at test time.

The second category of methods, which this proposal is also a member of, makes no explicit assumptions of what type of distribution shift occurs in the target domain. These methods include approaches such as \textit{unsupervised domain adaptation}~\cite{Hoffman2016wild,Hoffman_cycada2017}, \textit{train-on-test}~\cite{sun2019testtime,wang2020tent,hansen2020pad}, and \textit{tailoring}~\cite{alet2021tailoring}. These methods all tackle adaptation face-on, as an out-of-distribution generalization problem. Under this view, what happens in the target domain can not be known in advance when training the model. These approaches differentiate the adaptation phase from the training phase by what types of information is privileged, i.e., being only available during training. Examples include ground-truth labels under supervised learning, segmentation masks for dense predictions in computer vision, and instrumented rewards during reinforcement learning. Without these forms of privileged information available at test time, these approaches cast adaptation as an unsupervised, or self-supervised, learning process, with the main differences between methods being the learning objective, optimization details, and ways that they augment the data. In particular, test-time adaptation by entropy minimization~\cite{wang2020tent} shows that fine-tuning just the two parameters in layernorm gives better performance than fine-tuning the entire network. CycADA~\cite{Hoffman2017cycada} and FCN in-the-wild~\cite{Hoffman2016wild} use a cycle-consistent, adversarial loss for matching pixel-wise dense features. Some of these methods~\cite{Sankaranarayanan2017synthetic} produce feature alignment by synthesizing image pixels, whereas our proposal directly enforces distributional alignment in a compact latent space using an adversarial objective, without reconstructing image patches. This long line of work derive from classical unsupervised domain adaptation methods that pre-date adversarial generative techniques. \cite{Geng2011domain,Long2015adaptation} and~\cite{sun2019uda}, for instance, directly minimizes the measure of Maximum Mean Discrepancy (MMD) to great effect.

\section{Unsupervised Policy Adaptation}
\label{sec:problem}

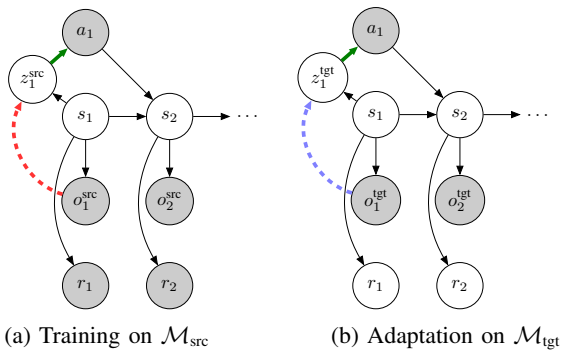
\begin{figure}[!t]
\centering
\begin{subfigure}[t]{0.24\textwidth}
\scalebox{0.7}{%
\usetikzlibrary{shapes.misc}
\begin{tikzpicture}[
  node distance=2em, auto,
  lat/.style={draw=black, circle, minimum size=2.5em},
  det/.style={draw=black, rectangle, minimum size=2.5em},
  obs/.style={circle, draw=black, fill=black!20, minimum size=2.5em},
  gen/.style={->, -{Stealth[length=.5em, inset=0pt]}},
  inf/.style={dashed, ->, -{Stealth[length=.5em, inset=0pt]}},
]
\node[obs, inner sep=.02em] (r1) {$r_1$};
\node[obs, right=of r1, inner sep=.02em] (r2) {$r_2$};
\node[obs, above=of r1] (o1) {$o^\textrm{src}_1$};
\node[obs, above=of r2] (o2) {$o^\textrm{src}_2$};
\node[lat, above=of o1] (x1) {$s_1$};
\node[lat, above=of o2] (x2) {$s_2$};
\node[right=of x2] (dots) {$\dots$};
\path (x2) edge[gen] node {} (dots);
\node at (x1) [lat, shift=(-3.5em :-3.5em)] (z1) {$z^\textrm{src}_1$};
\path (o1) edge[red!80, gen, bend left=50, line width=2, dashed] node{} (z1);
\node[obs, above=of x1] (a1) {$a_1$};
\path (z1) edge[darkgreen, gen, line width=2] node{} (a1);
\path (x1) edge[gen] node {} (z1);
\path (x1) edge[gen] node {} (o1);
\path (x1) edge[gen, bend right] node {} (r1);
\path (x2) edge[gen] node {} (o2);
\path (x2) edge[gen, bend right] node {} (r2);
\path (x1) edge[gen] node {} (x2);
\path (a1) edge[gen] node {} (x2);
\end{tikzpicture}}
\caption{Training on $\mathcal{M}_\textrm{src}$}\label{fig:source_mdp}
\end{subfigure}\hfill%
\begin{subfigure}[t]{0.24\textwidth}
\raisebox{0em}{
\scalebox{0.7}{%
\begin{tikzpicture}[
  node distance=2em, auto,
  lat/.style={draw=black, circle, minimum size=2.5em},
  det/.style={draw=black, rectangle, minimum size=2.5em},
  obs/.style={circle, draw=black, fill=black!20, minimum size=2.5em},
  gen/.style={->, -{Stealth[length=.5em, inset=0pt]}},
  inf/.style={dashed, ->, -{Stealth[length=.5em, inset=0pt]}},
]
\node[lat, inner sep=.02em] (r1) {$r_1$};
\node[lat, right=of r1, inner sep=.02em] (r2) {$r_2$};
\node[obs, above=of r1] (to1) {$o_1^\textrm{tgt}$};
\node[obs, right=of to1, above=of r2] (to2) {$o^\textrm{tgt}_2$};
\node[lat, above=of to1] (s1) {$s_1$};
\node[lat, above=of to2] (s2) {$s_2$};
\path (s1) edge[gen, bend right] node {} (r1);
\path (s2) edge[gen, bend right] node {} (r2);
\node[obs, above=of s1] (a1) {$a_1$};
\node[right=of s2] (dots) {$\dots$};
\path (s2) edge[gen] node {} (dots);
\node at (s1) [lat, shift=(-3.5em :-3.5em)] (z1) {$z^\textrm{tgt}_1$};
\path (to1) edge[blue!50, gen, bend left=50, line width=2, dashed] node{} (z1);
\path (s1) edge[gen] node {} (z1);
\path (a1) edge[gen] node {} (s2);
\path (s1) edge[gen] node {} (to1);
\path (s1) edge[gen] node {} (s2);
\path (s2) edge[gen] node {} (to2);
\path (z1) edge[darkgreen, gen, line width=2] node{} (a1);
\end{tikzpicture}}}
\caption{Adaptation on $\mathcal{M}_\textrm{tgt}$ }\label{fig:target_mdp}
\end{subfigure}%
\caption{
A Markov decision process gives rise to two different observations \(o^\src_t\) and \(o^\tgt_t\). \(o^\src_t\) is accessible during training, whereas \(o^\tgt_t\) is only accessible during deployment.
The {\color{red!80}red} dashed arrow indicates the learned inference network \(F\), produced as part of the policy during training. During deployment, the latent features produced by \(F\), \(z^\tgt_t\) ({\color{es-blue}blue} dashed arrow), experience a domain-shift. The reward further becomes unobservable. Our goal is to make the representation domain-invariant, such that the policy \(\pi(a\vert o^\tgt)\) (frozen policy head in  {\color{darkgreen}green}) can succeed in the target domain.}
\label{fig:graphicalmodels}
\end{figure}

Unsupervised policy adaptation is a setting that involves two distinct domains --- a source domain and a target domain. These two domains share the same underlying MDP and task structure, but has different observation conditions. In the target domain, the agent only has access to observations \(o_t^\text{tgt}\) and its own actions \(a_t\), but not the corresponding rewards or the ground-truth state \(s_t\) (see Figure~\ref{fig:graphicalmodels}). A practical example is a robot that is trained with images in a clean, simulated environment that now has to work in-the-wild, in the presence of visual distractors and changes in the lighting condition or the mounting pose of the video camera. These variations could lead to significantly different image observations. As a result of this shift, deploying an agent trained in the source domain directly in the target domain (i.e., zero-shot transfer) generally results in poor performance.

Formally, we consider an infinite horizon Markov decision process (MDP)~\citep{Puterman2014mdp}  \(\mathcal{M}\) parameterized via the tuple \(\langle S, A, O, R, P, \gamma\rangle\), where \(S\) and \(A\) are the state and action spaces. \(P:S\times A\mapsto S\) is the transition function, \(R: S \times A\mapsto \mathbb R\) is the scalar reward, and \(\gamma\) is the discount factor. The agent receives a stream of observations $o \in O$. We assume a fully-observable setting where a single observation carries enough information to decide an appropriate action. In the source domain \(\mathcal M_\text{src}\) (see Figure~\ref{fig:source_mdp}), we can use reinforcement learning to produce an optimal policy $\pi: O \times A \mapsto [0, 1]$ that maximizes the expected discounted return \(\mathcal J = \mathbb{E}\left[ \sum_\infty \gamma^t R(s_t, a_t)\right]\). In the target domain (see Figure~\ref{fig:target_mdp}) however, the reward is not observable therefore we can not rely on reinforcement learning for fine-tuning. Nevertheless the task structure remains identical to that of the source domain. 
We assume that the policy $\pi$ consists of an encoder $F: O \mapsto Z$, where $Z$ is a compact latent space, and a policy head $\pi_z:Z \times A \mapsto [0,1]$ shown as {\color{darkgreen}green} arrows. The goal of unsupervised policy adaptation is to find ways to battle this distribution shift, so that the resulting, adapted policy can succeed on the task in \(\mathcal M_\text{tgt}\).

\section{Invariance Through Latent Alignment}
\label{sec:ila}
\begin{figure*}[!t]
    \centering
    \includegraphics[width=\textwidth]{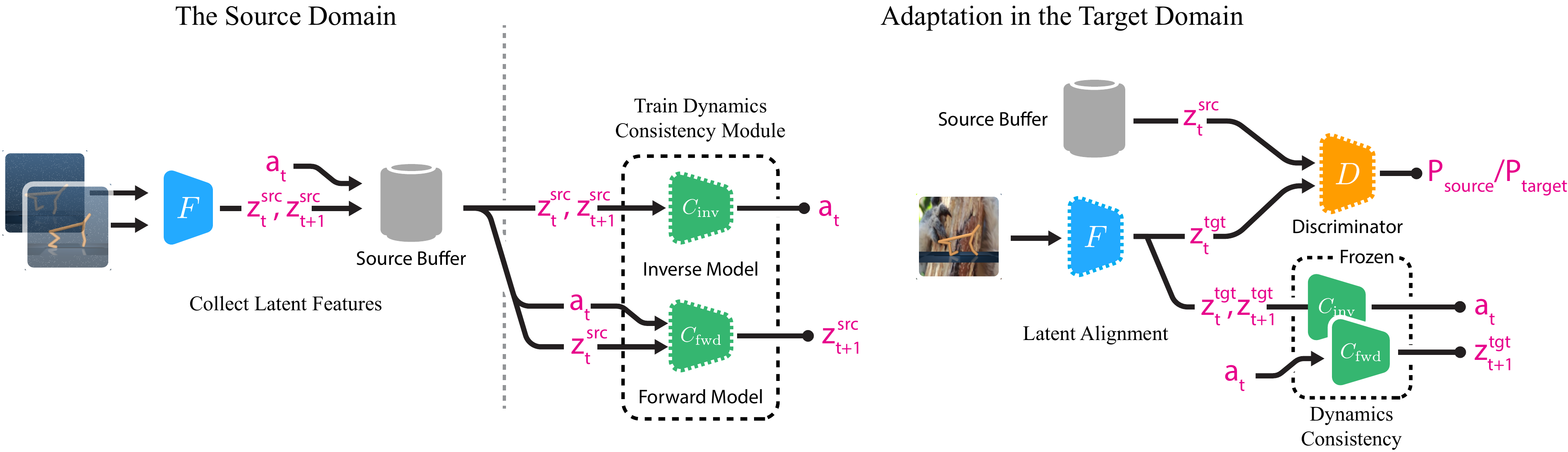}
    \caption{Training and adaptation phases of \lcila. The encoder $F$ takes an observation of target domain, and learns to fool the discriminator, while the discriminator $D$ predicts whether the input is an encoded target observation or a latent sample from source buffer. This adversarial training encourages the distribution of encoder outputs to be similar to the latent embedding sampled from the source buffer. $C_\textrm{fwd}$ and $C_\textrm{inv}$ are the forward and inverse dynamics networks that guides the encoder adaptation.}
    \label{fig:main-fig}
\end{figure*}

When we ask if the agent can perform in the target domain, we are effectively making the assumption that the task, and the underlying MDP has not changed. One way to factorize the problem is to divide the policy into two modules (see Figure~\ref{fig:source_mdp}). A \textit{policy head} \(\pi(a\vert z)\) that we keep \textit{frozen} during the adaptation process, and an \textit{encoder} \(F(z\vert o)\) that we adapt. Ideally, the latent feature \(z\) captures what the policy needs to know to accomplish the task. Hence we can formulate unsupervised policy adaptation as a distribution-matching objective that ``aligns'' the distribution \(P_\pi(z_\text{tgt})\), with the one in the source domain, \(P_\pi(z_\text{src})\) (see Figure~\ref{fig:latent-alignment}).

This overall \textit{sim-and-adaptation} pipeline starts in the source domain with collecting latent features \(z_\text{src}\) into a buffer. The agent carries these data into the deployment environment. Then during adaptation, it optimizes two objectives. The first is a minimax objective that focuses on individual latent features (\(D\) in Figure~\ref{fig:main-fig}). The second is an cooperative dynamics consistency objective that consists of both the forward and the inverse kinematics prediction error (\(C_\text{inv}\) and \(C_\text{fwd}\) in Figure~\ref{fig:main-fig}). The training procedure partially resembles a generative adversarial network (GAN)~\cite{Goodfellow2014gan} with two key distinctions. First, we do not reconstruct raw pixel observations but instead directly match the distribution of the latent features \(z\). This has the benefit that we do not require a generator that incur additional space and optimization overhead. Second, we find it helpful to pretrain the dynamics consistency module at the beginning of the adaptation process because it reduces the wall-clock time of the procedure.

We cover details of the procedure below.

\subsection{Collecting Latent Features in The Source Domain}

In the {\color{purple}source domain} we collect latent vectors \(z_t^\text{src}\) and actions \(a_t\) into a buffer
\begin{equation}
\mathcal B_\text{\src} = \{z_0^\text{src}, a_0; z_1^\text{src}, a_1,\dots\}\text{\quad where\quad}  z = F(o_t^\text{src}). 
\end{equation}The trajectories, \(\tau_\text{src} = \{o_0, a_0; o_1, a_1, \dots\}\) are sampled from \(\mathcal M_\text{src}\) with an exploration policy \(\pi\) that can be different from the pretraind policy for the task. In our experiment we found that using a random policy \(\bar \pi\) is unexpectedly effective, with the added benefit that the same policy can be used in the target domain.

\subsection{Dynamics Consistency}

To match the joint distribution \(P(z_t^\text{tgt}, a_t, z_{t+1}^\text{tgt})\) with those on the source domain, we introduce a dynamics consistency loss which is the sum of the \(\ell^2\) error in the forward and inverse dynamics predictions
\begin{equation} \label{eqn:dynamics_loss}
    \begin{split}
        \mathcal{L}_\textrm{dyn}(z_t, z_{t+1}, a_t) &= \lVert C_{\text{fwd}}(z_t, a_t) - z_{t+1} \rVert^2\\%
        &+ \lVert C_{\text{inv}}(z_t, z_{t+1}) - a_t \rVert^2.
    \end{split}
\end{equation}
\(C_\text{fwd}(z_{t+1} \vert  z_t, a_t)\) is the forward kinematics model that predicts the next latent $z_{t+1}$ given the previous latent $z_t$ and $a_t$. \(C_\text{inv}(a_t \vert z_t, z_{t+1})\) is the inverse model that predicts the action $a_t$ associated with the transition from $z_t$ to $z_{t+1}$. We found it was not necessary to scale the two terms separately as it worked well enough.

\begin{algorithm}[t]
\caption{Populating The Source Buffer}\label{alg:buffer}
\begin{algorithmic}[1]
\small
\Require Encoder $F$, empty buffer $\mathcal{B_\csrc}$, random policy $\bar{\pi}$
\For{step in $1,\dots, N$} \Comment{Collect latent features}
\State \text{Sample} $o_t, a_t, o_{t+1} \sim P_{\bar{\pi}}(\mathcal{M}_\csrc)$
\State \text{Encode} $z_t, z_{t+1} \gets F(o_t), F(o_{t+1})$
\State $\mathcal{B}_\csrc \gets \mathcal{B}_\csrc \cup (z_t, a_t, z_{t+1})$
\EndFor
\end{algorithmic}
\end{algorithm}
We optimize the parameters of \(C_\textrm{fwd}\), \(C_\textrm{inv}\) and \(F\) in a \textit{cooperative} manner as opposed to an adversarial one. \(C_\textrm{fwd}\) and \(C_\textrm{inv}\) are optimized using latent transitions sampled from the source buffer $\mathcal{B}_\text{src}$
\begin{equation}\label{eqn:dynamics-pretrain} 
C_\textrm{fwd}, C_\textrm{inv} = \argmin_{C_\textrm{fwd}, C_\textrm{inv}} \; 
\mathbb{E}_{z_t,a_t,z_{t+1}\sim \mathcal B_\csrc} \big[ \nonumber 
\mathcal{L}_\textrm{dyn}(\srceq{z^{\src}_t}, \srceq{z^{\src}_{t+1}}, \srceq{a^{\src}_t}) \big].
\end{equation}
To update the encoder \(F\) we sample transitions using a random policy \(\bar \pi\) from the {\color{blue!60}target domain}. We freeze the parameters of the two dynamics model when updating \(F\).
\begin{equation}
    \mathcal{J}_\textrm{dyn} =  \mathbb{E}_{o_t, a_t, o_{t+1}\sim P_{\bar{\pi}}(\mathcal{M}_\ctgt)} \left[ \mathcal{L}_\textrm{dyn}(F(\tgteq{o_{t}^\tgt}), F(\tgteq{o_{t+1}^\tgt}), \tgteq{a^\tgt_t})\right].
    \label{eqn:dynamics_consistency_loss}
\end{equation}
We found that a learning rate of \(1e^{-6}\) worked sufficiently well for both, and we did not find it necessary to scale the two loss terms separately.

\subsection{Adversarial Loss}

In addition to the dynamics consistency loss, we also introduce an adversarial learning objective (see Figure~\ref{fig:main-fig}), where a discriminator $D$ tries to distinguish between embeddings from the source domain $\srceq{z^{\src}_t}$ and those from the target domain $\tgteq{z^{\tgt}_t}$. We update the parameters of the encoder such that latent embeddings on the target domain are indistinguishable from those of the source domain.
Using the earth-moving metric from~\citep{arjovsky2017wasserstein}, we express this distribution-matching objective as
\begin{equation}\label{eqn:adversarial_loss}
        \mathcal{J}_\textrm{adv} = \mathbb{E}_{z\sim \mathcal B_\csrc} \bigl[ D\left(\srceq{z_t^\src})\right) \bigr]
      + \mathbb{E}_{P_{\bar{\pi}}(\mathcal{M}_\ctgt) }\bigl[1 - D\left(F(\tgteq{o_t^\tgt})\right) \bigr].
\end{equation}
The encoder tries to minimize this objective while the discriminator acts as an adversary and seeks to maximize it, resulting in a GAN-like minimax game.

\subsection{Putting Things Together}

\begin{algorithm}[t]
\caption{\ila}\label{alg:cap}
\begin{algorithmic}[1]
\small
\Require Pretrained encoder $F$, discriminator $D$, populated buffer $\mathcal{B_\csrc}$, $C_\textrm{fwd}, C_\textrm{inv}$, random policy $\bar{\pi}$

\For{step in $1,\dots T_\text{dyn}$}  \Comment{Pretrain dynamics networks}
\State \text{Sample} $z_t, a_t, z_{t+1} \sim \mathcal{B}_\csrc$
\State $\Delta_{C_\textrm{fwd}}, \Delta_{C_\textrm{inv}} \gets \nabla_{C_\textrm{fwd}, C_{\textrm{inv}}}\mathcal{L}_\textrm{dyn}(z_t, z_{t+1}, a_t)$
\State $C_\text{fwd}, C_\text{inv} \gets \text{Optim.step}(C_{\textrm{fwd}}, C_{\textrm{inv}}, \Delta_{C_\textrm{fwd}}, \Delta_{C_\textrm{inv}})$
\EndFor

\For{step in $1,\dots T$}  \Comment{Adaptation main loop}
\State \text{Sample} $\srceq{z^\src_t}, \srceq{a^\src_t}, \srceq{z^\src_{t+1}} \sim \mathcal{B}_\csrc$
\State \text{Sample} $\tgteq{o^\tgt_t}, \tgteq{a^\tgt_t}, \tgteq{o^\tgt_{t+1}} \sim \mathcal{P}_{\bar{\pi}}(\mathcal{M}_\ctgt)$
\State \text{Compute gradients}:
\State $\Delta_D \gets \nabla_{D} \big[ D(\srceq{z_t^{\src}}) + (1 - D(F(\tgteq{o_t^{\tgt}}))) \big]$ \Comment{Discriminator}
\State $\Delta_{F_1} \gets \nabla_{F} \big[ D(\srceq{z_t^{\src}}) + (1 - D(F(\tgteq{o_t^{\tgt}}))) \big]$
\State $\Delta_{F_2} \gets \nabla_{F} \mathcal{L}_\textrm{dyn}(F(\tgteq{o_t^{\tgt}}), F(\tgteq{o_{t+1}^{\tgt}}), \tgteq{a^\tgt_t})$ \Comment{Dyn. consistency}
\State $D \gets \text{Optim.step}(D, - \Delta_D)$
\State $F \gets \text{Optim.step}(F, \Delta_{F_1} + \Delta_{F_2})$
\EndFor
\end{algorithmic}
\label{alg:adaptation}
\end{algorithm}

\begin{figure*}[t]
    \begin{minipage}{0.34\textwidth}
    \includegraphics[width=\linewidth]{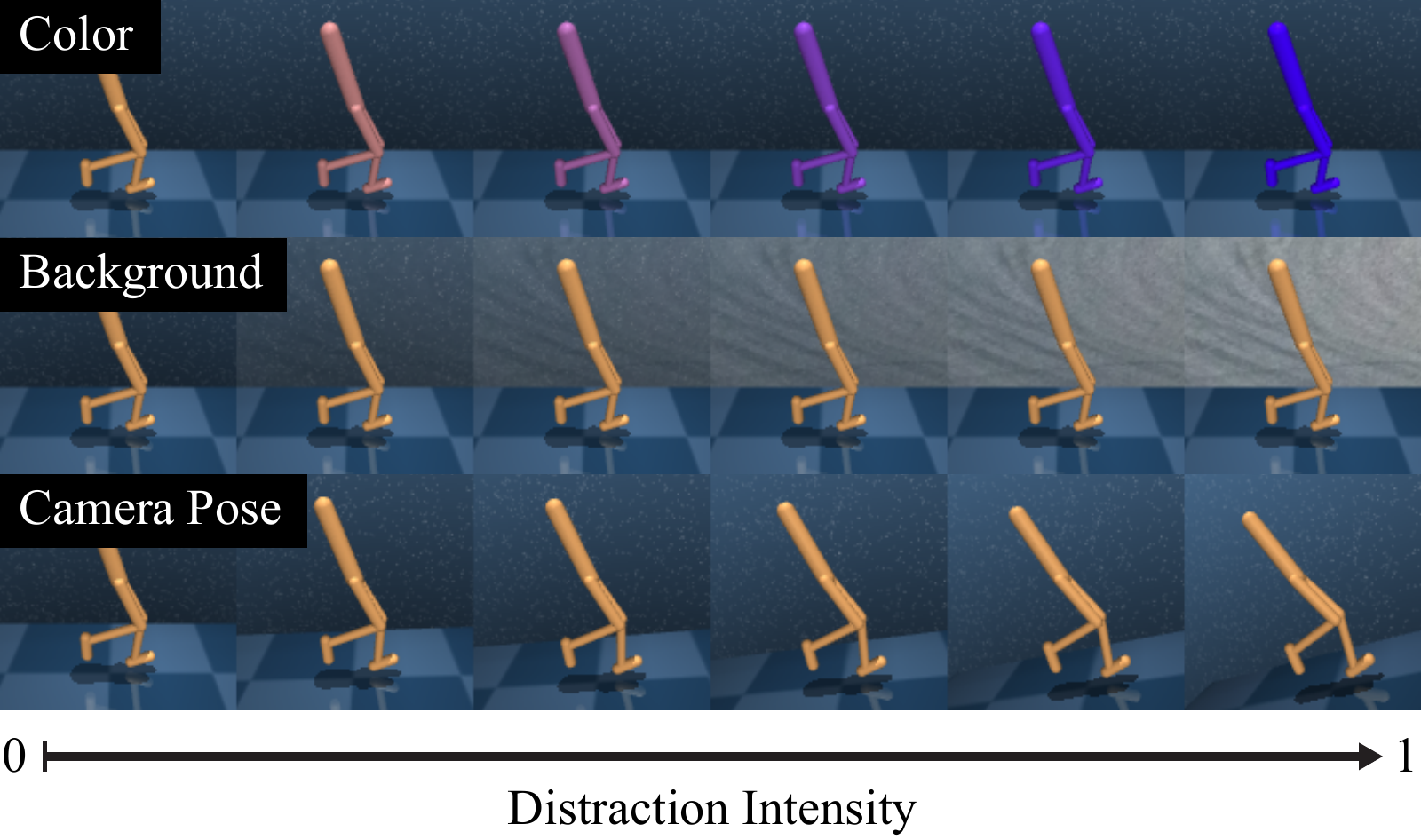}
    \caption{Samples from the modified \textit{distracting control suite}. Top row: color variations, middle row: background distractions, bottom row: camera pose variations.}
    \label{fig:distractions}
    \end{minipage}\hfill%
    \begin{minipage}{0.63\textwidth}
  \begin{subfigure}[t]{0.3\linewidth}
    \centering
    \includegraphics[height=80px]{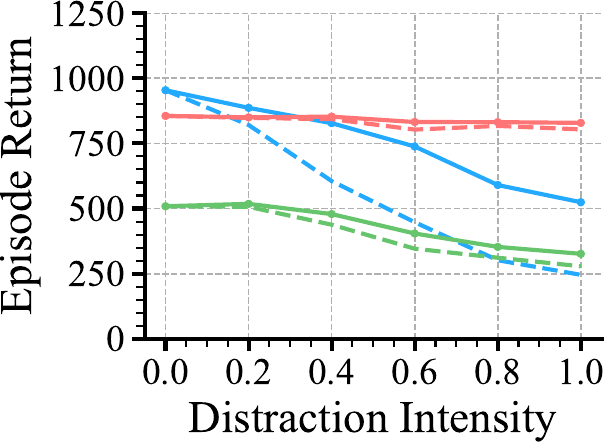}
    \caption{Color}
    \label{fig:intensity-colors}
  \end{subfigure}\hfill
  \begin{subfigure}[t]{0.3\linewidth}
    \centering
    \includegraphics[height=80px]{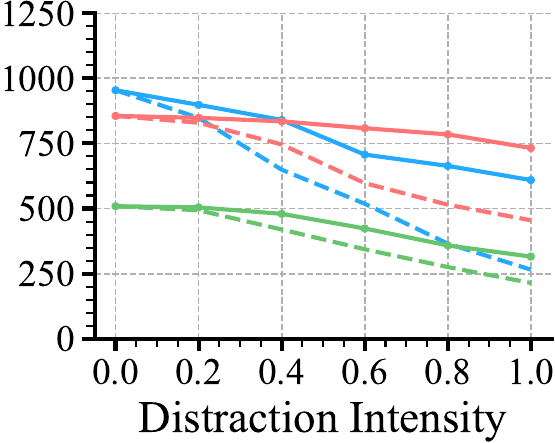}
    \caption{Background}
    \label{fig:intensity-background}
  \end{subfigure}\hfill
  \begin{subfigure}[t]{0.3\linewidth}
    \centering
    \includegraphics[height=80px]{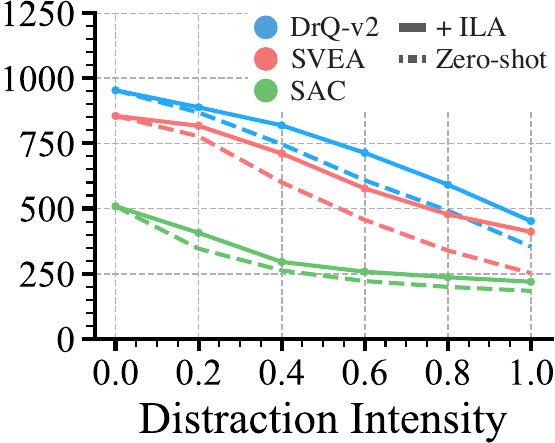}
    \caption{Camera Pose}
    \label{fig:intensity-camera}
  \end{subfigure}
    \vspace{0.5em}
    \caption{The gain of applying \ac{ILA} to target domains with various distraction intensities. Dashed lines denote the performance of the baseline agent in the target environment (i.e., zero-shot transfer), while solid lines represent the performance gains of the agents with ILA.}
    \label{fig:intensity-vs-reward}
  \end{minipage}
\end{figure*}

We adapt our encoder by minimizing a loss that combines both the adversarial loss $\mathcal{J}_\textrm{adv}$ (Eqn.~\ref{eqn:adversarial_loss}) and the dynamics consistency loss $\mathcal{J}_\textrm{dyn}$ (Eqn.~\ref{eqn:dynamics_consistency_loss}). Specifically, we solve for the parameters of the encoder through the following objective
\begin{equation}\label{eqn:joint_objective}
    \min_{F}\left[\max_D \mathcal{J}_\textrm{adv} + \mathcal{J}_\textrm{dyn}\right].
\end{equation}
We did not find it necessary to add additional scaling factors to balance the loss terms. 
Algorithm \ref{alg:buffer} and \ref{alg:adaptation} summarizes the whole procedure. We train the dynamics consistency networks to convergence before running the adaptation loss, to improve the wall-time.

The proposed approach, invariance through latent alignment, does not affect the training procedure of the control agent in the source domain, so it can be applied to any pretrained agent. This is also an unsupervised adaptation procedure as it does not require reward supervision at test time, nor does it require paired source and target images.

\section{Experiments}


\label{sec:experiments}


\begin{table*}[h!]
\centering
\caption{Episode return in the target (test) environments (mean and standard deviation) before (zero-shot) and after (+ILA) adaptation for SAC, SVEA, and DrQ-v2 with background distraction at an intensity setting of $1.0$. The performance of each baseline in the source (training) environments can be found in the Appendix.}\label{table:baseline-adaptation}{\small%
\setlength{\tabcolsep}{8pt}
\begin{tabularx}{0.8\linewidth}{Xcc|cc|cc}
    \toprule
                       & \multicolumn{2}{c}{SAC}                                          & \multicolumn{2}{c}{SVEA}                                                       & \multicolumn{2}{c}{DrQ-v2}\\
    Domain              & Zero-shot                 & +ILA                                  & Zero-shot                             & +ILA                                  & Zero-shot                     & +ILA \\
    \midrule
\texttt{ball\_in\_cup-catch} & $115^{\scriptscriptstyle\pm 50} \hphantom{0}$  & $\bm{227^{\scriptscriptstyle\pm 222}}$              & $490^{\scriptscriptstyle\pm 376}$                          & $\bm{987^{\scriptscriptstyle\pm 27}} \hphantom{0}$    & $\hphantom{0}88^{\scriptscriptstyle\pm 39} \hphantom{0}$     & $\bm{386^{\scriptscriptstyle\pm 425}}$    \\
\texttt{cartpole-balance}    & $434^{\scriptscriptstyle\pm 275}$             & $\bm{585^{\scriptscriptstyle\pm 295}}$              & $446^{\scriptscriptstyle\pm 330}$                          & $\bm{627^{\scriptscriptstyle\pm 258}}$               & $273^{\scriptscriptstyle\pm 107}$                           & $\bm{322^{\scriptscriptstyle\pm 117}}$    \\
\texttt{cartpole-swingup}    & $182^{\scriptscriptstyle\pm 147}$             & $\bm{369^{\scriptscriptstyle\pm 243}}$              & $269^{\scriptscriptstyle\pm 365}$                          & $\bm{612^{\scriptscriptstyle\pm 213}}$               & $\hphantom{0}82^{\scriptscriptstyle\pm 35}\hphantom{0}$     & $\bm{247^{\scriptscriptstyle\pm 136}}$    \\
\texttt{cheetah-run}         & $169^{\scriptscriptstyle\pm 65} \hphantom{0}$  & $\bm{248^{\scriptscriptstyle\pm 53}}\hphantom{0}$   & $317^{\scriptscriptstyle\pm 137}$                          &  $\bm{378^{\scriptscriptstyle\pm 55}}\hphantom{0}$    & $100^{\scriptscriptstyle\pm 88}\hphantom{0}$                & $\bm{393^{\scriptscriptstyle\pm 125}}$    \\
\texttt{finger-spin}         & $113^{\scriptscriptstyle\pm 162}$             & $\bm{192^{\scriptscriptstyle\pm 196}}$              & $391^{\scriptscriptstyle\pm 467}$                          & $\bm{943^{\scriptscriptstyle\pm 54}}\hphantom{0}$    & $207^{\scriptscriptstyle\pm 328}$                           & $\bm{769^{\scriptscriptstyle\pm 206}}$    \\
\texttt{finger-turn\_easy}   & $\bm{163^{\scriptscriptstyle\pm 99}}\hphantom{0}$  & $146^{\scriptscriptstyle\pm 33} \hphantom{0}$   & $278^{\scriptscriptstyle\pm 180}$                          & $\bm{491^{\scriptscriptstyle\pm 343}}$  &  $268^{\scriptscriptstyle\pm 241}$   & $\bm{914^{\scriptscriptstyle\pm 44}}\hphantom{0}$  \\
\texttt{reacher-easy}        & $179^{\scriptscriptstyle\pm 65}\hphantom{0}$  & $\bm{381^{\scriptscriptstyle\pm 76}}\hphantom{0}$   & $\hphantom{0}75^{\scriptscriptstyle\pm 77} \hphantom{0}$    & $\bm{624^{\scriptscriptstyle\pm 305}}$               & $\hphantom{0}58^{\scriptscriptstyle\pm 32}\hphantom{0}$     & $\bm{685^{\scriptscriptstyle\pm 211}}$    \\
\texttt{walker-stand}        & $330^{\scriptscriptstyle\pm 118}$             & $\bm{364^{\scriptscriptstyle\pm 115}}$              & $917^{\scriptscriptstyle\pm 138}$                          & $\bm{999^{\scriptscriptstyle\pm 12}} \hphantom{0}$    & $630^{\scriptscriptstyle\pm 197}$                           & $\bm{868^{\scriptscriptstyle\pm 151}}$    \\
\texttt{walker-walk}         & $242^{\scriptscriptstyle\pm 142}$             & $\bm{291^{\scriptscriptstyle\pm 134}}$              & $866^{\scriptscriptstyle\pm 45}\hphantom{0}$               & $\bm{924^{\scriptscriptstyle\pm 45}}\hphantom{0}$    & $326^{\scriptscriptstyle\pm 195}$                           & $\bm{770^{\scriptscriptstyle\pm 140}}$ \\  
    \bottomrule
\end{tabularx}}
\end{table*}

We want to understand the impact of test-time adaptation on an agent's ability to generalize out-of-distribution. This section will compare \ac{ILA} with two state-of-the-art reinforcement learning baselines that uses data-augmentation: SVEA~\citep{svea} and DrQ-v2~\citep{yarats2021drqv2}. 
Recall from the introduction that these methods vie for increased generalization capabilities by expanding the support of the training distribution. We expect the invariance produced this way to be less performant than unsupervised adaptation at test time that only needs to focus on one specific instance of perceptual variation. In our experiments, we will also compare against \textit{policy adaptation during deployment} (PAD, see~\citep{hansen2020pad}), a baseline that, like our method, adapts the policy without access to the reward at test time. 

To further probe the generalization abilities of test-time adaptation, we conduct an experiment where we vary the intensity of environmental distractions. The results show that test-time adaptaion significantly increases the policy performance during deployment. We will conclude with some general discussion and remarks regarding the design tradeoffs involved in test-time adaptation.

\paragraph{Setup} We conduct experiments on nine domains from the DeepMind Control Suite (DMC, see~\citep{deepmindcontrolsuite2018}) and treat it as the \textit{source} domain for training the RL agents. We use the Distracting Control Suite ~\citep{stone2021distracting} as the \textit{target} domain. Distracting control suite adds three types of distractions to DMC, including image background, random color texture, and changes to the camera pose. The intensity of these modes of distraction are calibrated. For details, refer to the accompanying report (see~\citealt{stone2021distracting})

\paragraph{Modifications to Distracting Control Suite}
The default configuration of distracting control suite changes distractions at the start of every episode (e.g., different background images are used at every episode). However, we are interested in measuring an agent's ability to perform adaptation across several episodes on the same target environment. Thus, we modify distracting control suite to sample a distraction once in the beginning of learning, and then use the same distraction across all learning epochs. This also ensures consistent evaluation across algorithms. In accordance with this change, we also modify the intensity benchmark from distracting control suite. In our experiments, intensity measures the deviation between an environment distraction and the train environment's default value. For example, intensity may measure how far the distracting color is from the default. Finally, we modify the environments to only apply a single distraction during testing (rather than all three) in order to better understand the impact of each type of distraction on overall performance. Figure~\ref{fig:distractions} shows an example of distractions across intensities on Walker-walk domain.

Table~\ref{table:baseline-adaptation} presents the results for the different distracting control suite domains in the presence of background distractions with an intensity level of $1.0$. Specifically, we compare the test-time performance of SAC, SVEA, and DrQ-v2 in each domain with the episode rewards that we achieve when using \ac{ILA} to adapt the encoder. The baseline algorithms employ image augmentation, which provides some robustness to variations at test time. Even then, however, we find that \ac{ILA} improves the test-time generalization of all three baseline policies in most domains, often resulting in significant performance gains. In cases where \ac{ILA} does not improve performance, the resulting reward is comparable to the baseline policy, i.e., \ac{ILA} does not result in a performance degradation.

Figure~\ref{fig:intensity-vs-reward} visualizes the performance of the different methods, averaged over the set of distracting control suite domains, as a function of the intensity of the distractions. Since the baseline methods are trained with image augmentation, they do exhibit some robustness to distraction. However, we see this robustness rapidly diminishes as the distraction intensity increases. In particular, large changes to camera pose or the image background proved challenging for standard augmentation procedures. Comparatively, \ac{ILA} makes it much smoother and slower degradation of performance. This supports our hypothesis that adaptation powered by unsupervised learning can significantly widen the generalization abilities of learning algorithms.


\subsection{ILA on DeepMind Control Suite}

%


This section studies the impact of test-time adaptation on the DeepMind control suite. We begin by pretraining soft actor-critic (SAC)~\citep{sacapps}, SVEA, and DrQ-v2 in a non-distracting training-time environment. After training, we evaluate the learned policies on test environments with distractions of various intensities. This evaluation is zero-shot, i.e., there is no additional training in the test environment.

\begin{figure}[b!]
    \begin{subfigure}[t]{0.35\linewidth}
        \includegraphics[height=70px]{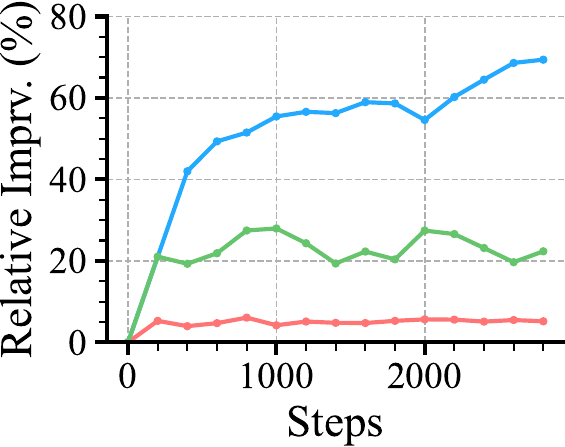}
        \caption{Color}
    \end{subfigure}\hfill
    \begin{subfigure}[t]{0.3\linewidth}
        \includegraphics[height=67.5px]{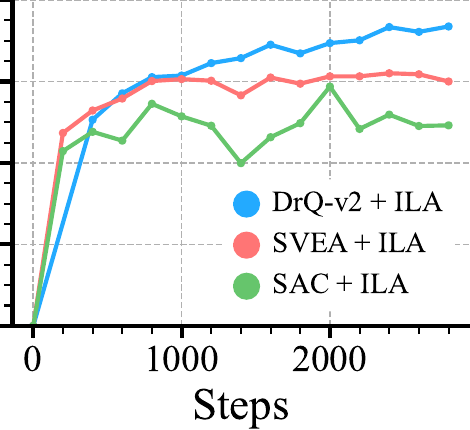}
        \caption{Background}
  \end{subfigure}\hfill
  \begin{subfigure}[t]{0.3\linewidth}
    \includegraphics[height=67.5px]{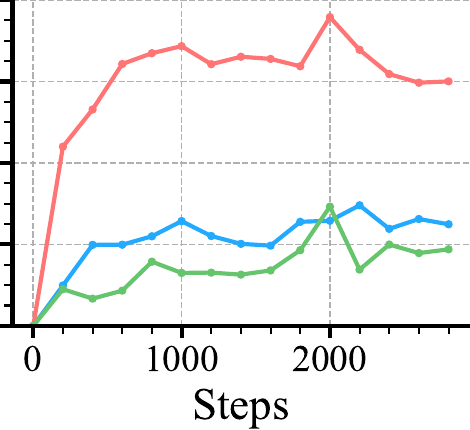}
    \caption{Camera Pose}
  \end{subfigure}\hfill
    \caption{Relative improvement (compared to zero-shot) as a function of adaptation steps when applying ILA to different baseline policies. As in Figure~\ref{fig:intensity-vs-reward}, each point represents the mean over nine domains and five random seeds. The results correspond to a distraction intensity value of $1$.}
    \label{fig:adaptation-curves}
\end{figure}


\subsection{Comparisons with PAD}
Similar to our approach, PAD pretrains the agent in a clean environment, and then adapts the agent via unsupervised objectives without assuming access to the target environment's reward function~\citep{hansen2020pad}. To evaluate the robustness of PAD to distractions, we consider distracting control suite with a fixed distraction intensity of $1.0$. Table~\ref{table:pad} compares the performance as the difference between the episode returns before and after adaptation along with the episode returns in the clean environment. It should be noted that PAD requires the policy to be trained along with an inverse dynamics prediction objective, whereas ILA does not. We include this additional auxiliary objective with soft actor-critic specifically for this experiment.

Across all environments, we see that PAD struggles to adapt to distractions at test time. We suspect this instability is caused by the large deviations in the latent variable distribution as a result of changes in the target environment. In particular, we posit that the signal from PAD's inverse dynamics head does not encourage the latent train and test distributions to match, whereas in \ac{ILA}, it does.

\subsection{Sim-to-real Transfer}
\begin{figure}[!t]
    \centering
    \begin{subfigure}[t]{0.34\linewidth}
    \includegraphics[width=\linewidth]{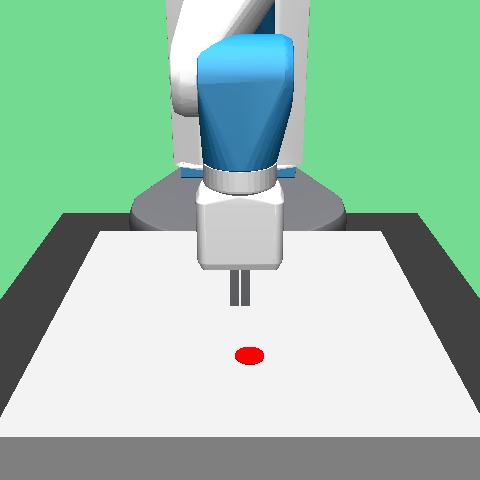}
    \end{subfigure}
    \hspace{2em}
    \begin{subfigure}[t]{0.34\linewidth}
    \includegraphics[width=\linewidth]{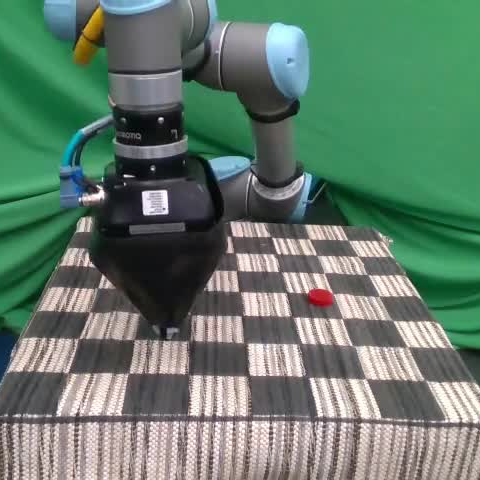}
    \end{subfigure}\hfill
    \caption{Simulated and real reach environment. The goal location is denoted as a red disk that the robot must reach. Using \ac{ILA}, we can transfer a policy trained in simulation (left) onto a real UR-5 robot (right). This adaptation requires no paired data and no rewards on the deployment environment, instead employing \ac{ILA} for unpaired adaptation.}\label{fig:sim2real}
\end{figure}

We are interested in the ability of \ac{ILA} to bridge the gap between simulated and real world robotics environments. In the {\it reaching} task, the target position is given by a red disc placed on a table. The agent's objective is to controls the arm so that the end-effector reaches this target location. Our goal is to train a policy in simulation, and then transfer the policy to a real UR-5 robot at test time. The same as with previous experiments, the test time agent receives no rewards. In both simulation and the real world, the policy's only input is an image from a camera placed in front of the robot and table. The action space is a 2D position controller that drives a small movement $(\delta x, \delta y)$ of the robotic gripper. See Figure~\ref{fig:sim2real} for the setup. 

We carry out the experiment by first training a policy with SVEA~\cite{svea} in simulation. For adaptation, we collect random trajectories in the real environment and then use \ac{ILA} to align the simulated and real world experiences. We evaluated the success rate of zero-shot and the adapted policies over \(20\) real episodes. We consider the episode to be success if the gripper's tip overlaps with the target in the front-view image.

Due to the challenging domain shift between simulator and real world, the zero-shot policy fails to adapt adequately, repeating the same action of moving the gripper to an edge of the table ad infinitum, regardless of the given goal location. This results in a final success rate of \(15\%\). 
On the same task, \ac{ILA} is able to robustly against this domain shift, achieving a final success rate around \(90\%\).


\subsection{Further Discussion}

\paragraph{Ablation Studies} In order to better understand the contribution of the different objectives to test-time generalization, we perform a series of ablations in which we omit either the dynamics consistency or the adversarial objectives.
%
%
In these experiments, we use a pretrained DrQ-v2 network for the algorithm's base policy, and then perform adaptation across all distractions with an intensity value of $1.0$. 
The results in Table \ref{table:ablation} show that the adversarial training is critical to adapt the latent representation in the target domain. Performing adaptation using only the dynamics consistency objective, i.e., $\argmin_F$ $\mathcal{J}_\textrm{dyn}$ (Eqn.~\ref{eqn:dynamics_consistency_loss}) results in a significant decrease in performance.
We theorize that the dynamics consistency objective helps to align latent transition manifolds when the latent distributions in source and target domains are reasonably close. 
If the latent distributions significantly differ, however, the input to the pretrained dynamics networks is largely out-of-distribution, and thus the gradients from dynamics consistency loss may negatively affect convergence. 

\begin{table}[b!]
    \centering
    \caption{Ablations with variants of \ac{ILA} that remove inverse/forward dynamics, or the adversarial objectives. DrQ-v2 is used as a pretrained policy. We compute episodic returns from nine domains and five random seeds, and the results correspond to an intensity value of $1$.} \label{table:ablation}
    {\small \begin{tabularx}{\linewidth}{Xccccc}
    \toprule
                 &           &      & +ILA & +ILA \\
    Distraction  & Zero-shot & +ILA & w/o dyn. & w/o adv. \\
    \midrule
    Background & $228^{\scriptscriptstyle\pm 232}$ & $602^{\scriptscriptstyle\pm 300}$ & $615^{\scriptscriptstyle\pm 289}$ & $176 ^{\scriptscriptstyle\pm 221}$ \\
    Colors  & $234^{\scriptscriptstyle\pm 245}$ & $536^{\scriptscriptstyle\pm 320}$ & $534^{\scriptscriptstyle\pm 327}$ & $117 ^{\scriptscriptstyle\pm 96}\hphantom{0}$ \\
    Camera Pose & $345^{\scriptscriptstyle\pm 287}$ & $417^{\scriptscriptstyle\pm 284}$ & $407^{\scriptscriptstyle\pm 272}$ & $208^{\scriptscriptstyle\pm 235}$ \\
    \bottomrule
    \end{tabularx}}
\end{table}
%
%
Compared to the adversarial objective, ablating the dynamics consistency objective
has surprisingly little effect on test-time generalization. 
It may be that the transition manifold in latent spaces
are preserved despite the distractions, which then diminishes the net effect of the dynamics consistency objective. 
\paragraph{Pre-Filling the Replay Buffer}
We implicitly make the assumption that a behavior policy $\bar{\pi}$ is available that can be used to generate trajectory data on both the source and the target domain with similar state visitation, and transition probabilities. To achieve good performance with the adapted policy, such distribution should also cover important states with higher reward. To our surprise, a simple scheme where we pre-fill both the source and target buffer using the random policy works sufficiently well.




\section{Closing Remarks}

\begin{table}[t]
    \centering
    \caption{Comparison with PAD}
    \begin{tabularx}{0.8\linewidth}{Xcccc}
    \toprule
    Distraction & Zero-shot & +PAD & +ILA\\
    \midrule
    
    None          & $835^{\scriptscriptstyle\pm 230} $ & ---  & --- \\ 
    Background    & $213^{\scriptscriptstyle\pm 247} $ & $ 279^{\scriptscriptstyle\pm 271}$ & $ \bm{425^{\scriptscriptstyle\pm 292}}$ \\
    Colors        & $230^{\scriptscriptstyle\pm 263}$ & $ 271^{\scriptscriptstyle\pm 300}$ & $ \bm{402^{\scriptscriptstyle\pm 339}}$ \\
    Camera Pose   & $319^{\scriptscriptstyle\pm 265}$ & $ 326^{\scriptscriptstyle\pm 259}$ & $ \bm{412^{\scriptscriptstyle\pm 275}}$ \\
    \bottomrule
    \end{tabularx}
    \label{table:pad}
\end{table}

We introduced \textit{\lcila}, an unsupervised approach that matches the distribution of feature vectors in the latent space, to improve the test-time performance of a learned visuomotor control policy. Empirical results show that as discrepancies between the training and deployment environments become more intense, \lcila has a large competitive edge over alternatives such as data augmentation techniques. The problem of test-time adaptation in visual reinforcement learning using unsupervised test-time trajectories is relatively new, but has thus far shown great relevance and promise in robotics~\cite{hansen2020pad,hansen2021soda}, where a sim-to-real pipeline has been at the fore-front of recent progress~\cite{Kaufmann2018drone,Wu2019deformable,margonlis2022rapid}.

A problem that remains is how to sample trajectories for adaptation. Technically, the distribution of trajectories and their latent representation on the source domain depends on the exploration policy to collect data. There are state transitions that are only accessible through a performant policy. To our surprise, simply using the random policy to collect data on both the source and the target domain largely by-passes this issue and achieves good performance post-adaptation. Future work in this area might want to develop better techniques for data generation at test time, so that the collected trajectory data better resembles those collected by the agent in the source buffer. One version towards this direction resembles generative adversarial imitation learning (GAIL, see~\cite{Ho2016gail}). Another open problem is that the performance does not improve monotonically upon more unsupervised updates. Finding an improvement scheme that is provably monotonic would be interesting.

Finally, although the adversarial distribution matching objective proposed here produce measurable improvements, better ways to align latent features are still needed. For a recent work in this direction, we refer the reader to \textit{adversarial support alignment}~\cite{tong2022support}.

\section{Acknowledgements}

This work was funded in part by the National Science Foundation under grant IIS-1830660. Ge Yang is supported by the National Science Foundation Institute for Artificial Intelligence and Fundamental Interactions (IAIFI, \href{https://iaifi.org/}{https://iaifi.org/}) under the Cooperative Agreement PHY-2019786. The authors would like to thank Adam Bohlander and Greg Shakhnarovich at the Toyota Technological Institute at Chicago (TTIC) for computing resources and additional storage. This paper also benefited from thoughtful discussions with students in the TTIC robotics group. The authors would also like to acknowledge the MIT SuperCloud and Lincoln Laboratory Supercomputing Center for providing high performance computing resources.


\bibliography{references}
\bibliographystyle{plainnat}

\clearpage
\onecolumn
\appendix

\subsection{Architectures}

This section describes the architectures of encoder $F$, discriminator $D$, inverse dynamics $C_\textrm{inv}$ and forward dynamics $C_\textrm{fwd}$. Encoder architectures follow the originally presented design choices of each base policy except for DrQ-v2. We take the part of the original network that produces a latent for the actor, and use it as an encoder $F$. In all of SAC (of the version used in \citep{svea}), SVEA and PAD, this shared latent is set to have dimension $100$. For DrQ-v2, we took the entire network architecture from SVEA.
The discriminator consists of a linear layer with hidden dimension $100$ followed by Layer Normalization (LN) \citep{ba2016layer} and tanh activation, and a three layer multi-layer perceptron (MLP) with and ReLU activations.
Inverse dynamics network $C_\textrm{inv}$ is five layer MLP and ReLU activations. It takes the concatenated latents as its input. Forward dynamics network $C_\textrm{fwd}$ takes action and latent, and encode them separately followed by concatenation and further layers. The action is fed it to a linear layer followed by LN, another linear layer and ReLU with hidden dimension 100. The latent is fed to three layer MLP where the first activation is LN and others are ReLU. The both encoded  inputs are concatenated and fed to 4 layer MLP with ReLU activations followed by LN and tanh activation. In all layers except for those explicitly mentioned, hidden dimension is set to $1,024$.

\subsection{Hyperparameters}

This section details the hyperparameter settings that were used for the experimental evaluation.
Table~\ref{table:hyp-pretraining} lists the hyperparameters relevant to pretraining the dynamics networks, while Table \ref{table:hyp-adaptation} provides those relevant to adaptation.
For the base policies, we adopted the hyperparameter settings and architecture choices from the original papers. Details for SAC and SVEA can be found in \citet{svea}, while \citet{hansen2020deployment} provides settings for PAD. The dimension of the latent $z_t$ differs depending on the encoder of the base policy, but all of the encoders in our experiments produce a latent with dimension $100$.

\begin{table}[!ht]
\centering
{\footnotesize%
\setlength{\tabcolsep}{5pt}
\caption{Hyperparameters for dynamics pretraining}\label{table:hyp-pretraining}
\begin{tabularx}{0.6\linewidth}{Xl}
    \toprule
Hyperparameter              & Value     \\
\midrule
Steps ($T_\textrm{dyn}$) & $100,000$ \\
Batch size  & $256$ \\
Optimizer & RMSProp($\alpha=0.99, \epsilon=1.0 \times 10^{-8}$) \\
Learning rate (forward dynamics) & $0.001$ \\
Learning rate (inverse dynamics) & $0.001$ \\
\bottomrule
\end{tabularx}}
\end{table}

\begin{table}[!th]
\centering
{\footnotesize%
\setlength{\tabcolsep}{5pt}
\caption{Hyperparameters for adaptation} \label{table:hyp-adaptation}
\begin{tabularx}{0.6\linewidth}{Xl}
    \toprule
Hyperparameter              & Value     \\
\midrule
Capacity of buffers ($N_\textrm{buf}$)  &  $1,000,000$ \\
Batch size & $256$ \\
Discriminator updates per step & $5$ \\
Gradient clipping & $0.01$ \\
Optimizer & RMSProp($\alpha=0.99, \epsilon=1.0 \times 10^{-8}$)  \\
Learning rate (encoder)  & $1.0 \times 10^{-4}$ (for DrQ-v2) \\
                                            & $1.0 \times 10^{-5}$ (otherwise)\\
Learning rate (discriminator) & $1.0 \times 10^{-4}$ (for DrQ-v2) \\
                                              & $1.0 \times 10^{-5}$ (otherwise) \\
Learning rate (inverse dynamics) & $1.0 \times 10^{-6}$ \\
\bottomrule
\end{tabularx}}
\end{table}

\clearpage
\subsection{Performance in original domains}

Table~\ref{table:clean-performance} presents the average reward for the baseline SAC, SVEA, and DrQ-v2 on the non-distracted source domains. The method labeled SAC+Inv denotes a soft actor-critic agent that is trained using inverse dynamics as an additional auxiliary objective that is only used in making a comparison against PAD.

\begin{table}[ht]
\centering
{\small
\caption{Performance in the source (clean) domains. }\label{table:clean-performance}
\begin{tabularx}{0.6\linewidth}{Xcccc}
    \toprule
Domain              & SAC           & SAC+Inv & SVEA          & DrQ-v2        \\
\midrule
\texttt{ball\_in\_cup-catch} & $\hphantom{0}452^{\scriptscriptstyle\pm 303}$             & $999^{\scriptscriptstyle\pm 7}\hphantom{00}$  & $1007^{\scriptscriptstyle\pm 4}\hphantom{00}$             & $1007^{\scriptscriptstyle\pm 3}\hphantom{00}$  \\
\texttt{cartpole-balance}    & $1022^{\scriptscriptstyle\pm 8}\hphantom{00}$             & $988^{\scriptscriptstyle\pm 26}\hphantom{0}$  & $\hphantom{0}996^{\scriptscriptstyle\pm 21}\hphantom{0}$  & $\hphantom{0}969^{\scriptscriptstyle\pm 123}$ \\
\texttt{cartpole-swingup}    & $\hphantom{0}735^{\scriptscriptstyle\pm 167}$             & $885^{\scriptscriptstyle\pm 24}\hphantom{0}$  & $\hphantom{0}892^{\scriptscriptstyle\pm 15}\hphantom{0}$  & $\hphantom{0}874^{\scriptscriptstyle\pm 21}\hphantom{0}$  \\
\texttt{cheetah-run}         & $\hphantom{0}309^{\scriptscriptstyle\pm 26}\hphantom{0}$  & $415^{\scriptscriptstyle\pm 59}\hphantom{0}$  & $\hphantom{0}448^{\scriptscriptstyle\pm 105}$             & $\hphantom{0}897^{\scriptscriptstyle\pm 45}\hphantom{0}$  \\
\texttt{finger-spin}         & $\hphantom{0}615^{\scriptscriptstyle\pm 63}\hphantom{0}$  & $971^{\scriptscriptstyle\pm 64}\hphantom{0}$  & $1000^{\scriptscriptstyle\pm 37}\hphantom{0}$             & $\hphantom{0}997^{\scriptscriptstyle\pm 36}\hphantom{0}$  \\
\texttt{finger-turn\_easy}   & $\hphantom{0}138^{\scriptscriptstyle\pm 28}\hphantom{0}$  & $658^{\scriptscriptstyle\pm 141}$             & $\hphantom{0}539^{\scriptscriptstyle\pm 317}$             & $\hphantom{0}945^{\scriptscriptstyle\pm 46}\hphantom{0}$  \\
\texttt{reacher-easy}        & $\hphantom{0}381^{\scriptscriptstyle\pm 39}\hphantom{0}$  & $723^{\scriptscriptstyle\pm 383}$             & $\hphantom{0}812^{\scriptscriptstyle\pm 293}$             & $\hphantom{0}988^{\scriptscriptstyle\pm 28}\hphantom{0}$  \\
\texttt{walker-stand}        & $\hphantom{0}438^{\scriptscriptstyle\pm 111}$             & $997^{\scriptscriptstyle\pm 6}\hphantom{00}$  & $1006^{\scriptscriptstyle\pm 4}\hphantom{00}$             & $\hphantom{0}996^{\scriptscriptstyle\pm 31}\hphantom{0}$  \\
\texttt{walker-walk}         & $\hphantom{0}393^{\scriptscriptstyle\pm 117}$             & $915^{\scriptscriptstyle\pm 22}\hphantom{0}$  & $\hphantom{0}964^{\scriptscriptstyle\pm 34}\hphantom{0}$  & $\hphantom{0}980^{\scriptscriptstyle\pm 16}\hphantom{0}$  \\
\bottomrule
\end{tabularx}}
\end{table}

\subsection{Ablation Studies}

Tables~\ref{table:ablation-background},~\ref{table:ablation-color}, and \ref{table:ablation-camera-pose} provide a per-domain ablation summary for background, color, and camera pose distractions, respectively. As with the results in Table~\ref{table:ablation}, we use DrQ-v2 as the pretrained policy and present the mean reward and standard deviation for five random seeds.
\begin{table}[!ht]
\centering
{\small%
\setlength{\tabcolsep}{5pt}
\caption{Ablation with background distraction}\label{table:ablation-background}
\begin{tabularx}{0.8\linewidth}{Xccccccc}
    \toprule
    &   &    & +ILA  & +ILA & +ILA & +ILA   \\
Domain      & Zero-shot     & +ILA          & w/o inv., fwd.      & w/o inv.      & w/o fwd.      & w/o adv.      \\
\midrule
\texttt{walker-walk}         & $326^{\scriptscriptstyle\pm 196}$ & $749^{\scriptscriptstyle\pm 133}$ & $778^{\scriptscriptstyle\pm 142}$ & $777^{\scriptscriptstyle\pm 145}$ & $768^{\scriptscriptstyle\pm 146}$ & $268^{\scriptscriptstyle\pm 367}$ \\
\texttt{walker-stand}     & $623^{\scriptscriptstyle\pm 233}$ & $866^{\scriptscriptstyle\pm 153}$ & $883^{\scriptscriptstyle\pm 110}$ & $859^{\scriptscriptstyle\pm 144}$ & $873^{\scriptscriptstyle\pm 129}$ & $402^{\scriptscriptstyle\pm 347}$ \\
\texttt{cartpole-swingup}    & $82^{\scriptscriptstyle\pm 35}$   & $231^{\scriptscriptstyle\pm 128}$ & $395^{\scriptscriptstyle\pm 218}$ & $288^{\scriptscriptstyle\pm 163}$ & $246^{\scriptscriptstyle\pm 152}$ & $117^{\scriptscriptstyle\pm 43}\hphantom{0}$  \\
\texttt{ball\_in\_cup-catch} & $88^{\scriptscriptstyle\pm 39}$   & $394^{\scriptscriptstyle\pm 387}$ & $381^{\scriptscriptstyle\pm 416}$ & $400^{\scriptscriptstyle\pm 414}$ & $383^{\scriptscriptstyle\pm 381}$ & $\hphantom{0}93^{\scriptscriptstyle\pm 42}\hphantom{0}$   \\
\texttt{finger-spin}         & $208^{\scriptscriptstyle\pm 327}$ & $783^{\scriptscriptstyle\pm 214}$ & $742^{\scriptscriptstyle\pm 190}$ & $772^{\scriptscriptstyle\pm 225}$ & $769^{\scriptscriptstyle\pm 223}$ & $\hphantom{0}74^{\scriptscriptstyle\pm 160}$  \\
\texttt{reacher-easy}        & $98^{\scriptscriptstyle\pm 93}$   & $726^{\scriptscriptstyle\pm 149}$ & $713^{\scriptscriptstyle\pm 111}$ & $732^{\scriptscriptstyle\pm 104}$ & $694^{\scriptscriptstyle\pm 169}$ & $\hphantom{0}91^{\scriptscriptstyle\pm 54}\hphantom{0}$   \\
\texttt{cheetah-run}         & $98^{\scriptscriptstyle\pm 90}$   & $411^{\scriptscriptstyle\pm 191}$ & $397^{\scriptscriptstyle\pm 152}$ & $398^{\scriptscriptstyle\pm 147}$ & $419^{\scriptscriptstyle\pm 157}$ & $\hphantom{0}12^{\scriptscriptstyle\pm 19}\hphantom{0}$   \\
\texttt{cartpole-balance}    & $271^{\scriptscriptstyle\pm 101}$ & $336^{\scriptscriptstyle\pm 126}$ & $315^{\scriptscriptstyle\pm 98}\hphantom{0}$  & $367^{\scriptscriptstyle\pm 84}\hphantom{0}$  & $297^{\scriptscriptstyle\pm 121}$ & $264^{\scriptscriptstyle\pm 80}\hphantom{0}$  \\
\texttt{finger-turn\_easy}   & $261^{\scriptscriptstyle\pm 244}$ & $920^{\scriptscriptstyle\pm 41}\hphantom{0}$  & $929^{\scriptscriptstyle\pm 57}\hphantom{0}$  & $899^{\scriptscriptstyle\pm 52}\hphantom{0}$  & $909^{\scriptscriptstyle\pm 74}\hphantom{0}$  & $256^{\scriptscriptstyle\pm 264}$ \\
\bottomrule
\end{tabularx}}
\end{table}
\begin{table}[!ht]
\centering
{\small%
\setlength{\tabcolsep}{5pt}
\caption{Ablation with color distraction}\label{table:ablation-color}
\begin{tabularx}{0.8\linewidth}{Xccccccc}
    \toprule
   &   &    & +ILA  & +ILA & +ILA & +ILA   \\
Domain              & Zero-shot     & +ILA           & w/o inv., fwd.      & w/o inv.      & w/o fwd.      & w/o adv.      \\
\midrule
\texttt{walker-walk}         & $\hphantom{0}80^{\scriptscriptstyle\pm 43}\hphantom{0}$   & $481^{\scriptscriptstyle\pm 313}$ & $468^{\scriptscriptstyle\pm 305}$ & $495^{\scriptscriptstyle\pm 330}$ & $472^{\scriptscriptstyle\pm 319}$ & $\hphantom{0}26^{\scriptscriptstyle\pm 5}\hphantom{00}$    \\
\texttt{walker-stand}        & $278^{\scriptscriptstyle\pm 150}$                         & $543^{\scriptscriptstyle\pm 231}$ & $603^{\scriptscriptstyle\pm 283}$ & $571^{\scriptscriptstyle\pm 259}$ & $548^{\scriptscriptstyle\pm 268}$ & $145^{\scriptscriptstyle\pm 31}\hphantom{0}$  \\
\texttt{cartpole-swingup}    & $152^{\scriptscriptstyle\pm 84}\hphantom{0}$              & $552^{\scriptscriptstyle\pm 377}$ & $520^{\scriptscriptstyle\pm 359}$ & $507^{\scriptscriptstyle\pm 339}$ & $434^{\scriptscriptstyle\pm 395}$ & $\hphantom{0}94^{\scriptscriptstyle\pm 56}\hphantom{0}$   \\
\texttt{ball\_in\_cup-catch} & $239^{\scriptscriptstyle\pm 374}$                         & $812^{\scriptscriptstyle\pm 225}$ & $857^{\scriptscriptstyle\pm 182}$ & $840^{\scriptscriptstyle\pm 251}$ & $843^{\scriptscriptstyle\pm 196}$ & $131^{\scriptscriptstyle\pm 46}\hphantom{0}$  \\
\texttt{finger-spin}         & $349^{\scriptscriptstyle\pm 354}$                         & $612^{\scriptscriptstyle\pm 345}$ & $561^{\scriptscriptstyle\pm 376}$ & $591^{\scriptscriptstyle\pm 351}$ & $603^{\scriptscriptstyle\pm 358}$ & $143^{\scriptscriptstyle\pm 172}$ \\
\texttt{reacher-easy}        & $138^{\scriptscriptstyle\pm 135}$                         & $490^{\scriptscriptstyle\pm 416}$ & $486^{\scriptscriptstyle\pm 392}$ & $486^{\scriptscriptstyle\pm 362}$ & $445^{\scriptscriptstyle\pm 405}$ & $140^{\scriptscriptstyle\pm 90}\hphantom{0}$  \\
\texttt{cheetah-run}         & $193^{\scriptscriptstyle\pm 188}$                         & $422^{\scriptscriptstyle\pm 282}$ & $416^{\scriptscriptstyle\pm 293}$ & $421^{\scriptscriptstyle\pm 277}$ & $406^{\scriptscriptstyle\pm 285}$ & $\hphantom{00}4^{\scriptscriptstyle\pm 2}\hphantom{00}$     \\
\texttt{cartpole-balance}    & $481^{\scriptscriptstyle\pm 351}$                         & $602^{\scriptscriptstyle\pm 366}$ & $576^{\scriptscriptstyle\pm 359}$ & $593^{\scriptscriptstyle\pm 351}$ & $583^{\scriptscriptstyle\pm 349}$ & $216^{\scriptscriptstyle\pm 94}\hphantom{0}$  \\
\texttt{finger-turn\_easy}   & $194^{\scriptscriptstyle\pm 200}$                         & $313^{\scriptscriptstyle\pm 290}$ & $323^{\scriptscriptstyle\pm 337}$ & $297^{\scriptscriptstyle\pm 316}$ & $304^{\scriptscriptstyle\pm 323}$ & $170^{\scriptscriptstyle\pm 57}\hphantom{0}$  \\
\bottomrule
\end{tabularx}}
\end{table}
\begin{table}[!ht]
\centering
{\small%
\setlength{\tabcolsep}{5pt}
\caption{Ablation with camera pose distraction}\label{table:ablation-camera-pose}
\begin{tabularx}{0.8\linewidth}{Xccccccc}
    \toprule
  &   &    & +ILA  & +ILA & +ILA & +ILA   \\
Domain              & Zero-shot     & +ILA           & w/o inv., fwd.      & w/o inv.      & w/o fwd.      & w/o adv.      \\
\midrule
\texttt{walker-walk}         & $293^{\scriptscriptstyle\pm 195}$                         & $375^{\scriptscriptstyle\pm 154}$             & $369^{\scriptscriptstyle\pm 168}$             & $389^{\scriptscriptstyle\pm 161}$             & $366^{\scriptscriptstyle\pm 164}$             & $\hphantom{0}63^{\scriptscriptstyle\pm 71}\hphantom{0}$   \\
\texttt{walker-stand}        & $621^{\scriptscriptstyle\pm 202}$                         & $704^{\scriptscriptstyle\pm 105}$             & $617^{\scriptscriptstyle\pm 155}$             & $661^{\scriptscriptstyle\pm 146}$             & $679^{\scriptscriptstyle\pm 93}\hphantom{0}$  & $330^{\scriptscriptstyle\pm 252}$ \\
\texttt{cartpole-swingup}    & $286^{\scriptscriptstyle\pm 51}\hphantom{0}$              & $234^{\scriptscriptstyle\pm 123}$             & $211^{\scriptscriptstyle\pm 122}$             & $241^{\scriptscriptstyle\pm 140}$             & $281^{\scriptscriptstyle\pm 42}\hphantom{0}$  & $172^{\scriptscriptstyle\pm 132}$ \\
\texttt{ball\_in\_cup-catch} & $327^{\scriptscriptstyle\pm 227}$                         & $462^{\scriptscriptstyle\pm 307}$             & $429^{\scriptscriptstyle\pm 321}$             & $566^{\scriptscriptstyle\pm 314}$             & $400^{\scriptscriptstyle\pm 337}$             & $199^{\scriptscriptstyle\pm 169}$ \\
\texttt{finger-spin}         & $\hphantom{0}29^{\scriptscriptstyle\pm 25}\hphantom{0}$   & $252^{\scriptscriptstyle\pm 216}$             & $222^{\scriptscriptstyle\pm 209}$             & $275^{\scriptscriptstyle\pm 194}$             & $251^{\scriptscriptstyle\pm 206}$             & $\hphantom{0}32^{\scriptscriptstyle\pm 63}\hphantom{0}$   \\
\texttt{reacher-easy}        & $917^{\scriptscriptstyle\pm 101}$                         & $933^{\scriptscriptstyle\pm 47}\hphantom{0}$  & $925^{\scriptscriptstyle\pm 60}\hphantom{0}$  & $950^{\scriptscriptstyle\pm 72}\hphantom{0}$  & $940^{\scriptscriptstyle\pm 90}\hphantom{0}$  & $732^{\scriptscriptstyle\pm 159}$ \\
\texttt{cheetah-run}         & $\hphantom{0}55^{\scriptscriptstyle\pm 20}\hphantom{0}$   & $142^{\scriptscriptstyle\pm 57}\hphantom{0}$  & $154^{\scriptscriptstyle\pm 90}\hphantom{0}$  & $141^{\scriptscriptstyle\pm 74}\hphantom{0}$  & $144^{\scriptscriptstyle\pm 61}\hphantom{0}$  & $\hphantom{0}24^{\scriptscriptstyle\pm 28}\hphantom{0}$   \\
\texttt{cartpole-balance}    & $288^{\scriptscriptstyle\pm 46}\hphantom{0}$              & $307^{\scriptscriptstyle\pm 100}$             & $375^{\scriptscriptstyle\pm 83}\hphantom{0}$  & $379^{\scriptscriptstyle\pm 116}$             & $380^{\scriptscriptstyle\pm 64}\hphantom{0}$  & $217^{\scriptscriptstyle\pm 47}\hphantom{0}$  \\
\texttt{finger-turn\_easy}   & $287^{\scriptscriptstyle\pm 89}\hphantom{0}$              & $346^{\scriptscriptstyle\pm 219}$             & $359^{\scriptscriptstyle\pm 122}$             & $377^{\scriptscriptstyle\pm 200}$             & $421^{\scriptscriptstyle\pm 180}$             & $160^{\scriptscriptstyle\pm 97}\hphantom{0}$ \\
\bottomrule
\end{tabularx}}
\end{table}

\end{document}